\def\mydegrees{Bachelor of Advanced Computing}

\documentclass[]{style/usydthesis}

\newcommand*{\noaddvspace}{\renewcommand*{\addvspace}[1]{}}
\addtocontents{lof}{\protect\noaddvspace}
\usepackage{graphicx}
\usepackage{amstext,amssymb,amsfonts,latexsym}
\usepackage{tikz}
\usepackage[margin=1in]{geometry}
\usepackage[utf8]{inputenc}
\usepackage{amsmath}
\usepackage{pgfplots}
\pgfplotsset{compat=1.17}
\usepackage{style/bib}
\usepackage[UKenglish]{babel}
\usepackage{csquotes}
\usepackage{ccicons}
\usepackage{animate}
\usepackage{setspace}
\usepackage{pdfpages}
\usepackage{hyperref}
\hypersetup{
    pdfauthor = {Christian Fane},
    pdftitle = {A Real-Time Diminished Reality Approach to Privacy in AR Collaboration},
    colorlinks,
    linkcolor={black},
    citecolor={black}
    }

\usepackage{microtype}

\begin{document}

\renewcommand{\thepage}{\roman{page}}	
\title{{\bf\Huge A Real-Time Diminished Reality Approach to Privacy in MR Collaboration}}
\author{Christian Fane}

\maketitle
\setstretch{1.5}



\chapter*{Acknowledgements}

Many thanks to Eduardo Velloso for his calming guidance throughout the project. Thank you to Pat for celebrating each of the small wins over the past year with me and for being my experimental dummy when needed. Finally, thank you to my parents for providing me with a fantastic educational pathway that has led me to such exciting projects!

\setcounter{tocdepth}{2}
\newpage
\addcontentsline{toc}{chapter}{Contents}
\tableofcontents
\listoffigures


\setcounter{page}{1}
\setcounter{chapter}{0}

\renewcommand{\thepage}{\arabic{page}}	
\setupParagraphs

\chapter{Introduction}\label{cha:introduction}

Mixed reality (MR) technologies have advanced dramatically in recent years, with powerful headsets now being produced by a range of competing companies. As the technology becomes more mainstream and accessible to the public \cite{Hornsey2024May}, considerations of user privacy must be addressed. Appropriate privacy tools are essential to support the responsible growth of MR applications. Because these devices rely on the continuous capture of visual and spatial data from the user’s surroundings, ongoing efforts are needed to ensure that private information is adequately protected.

Privacy concerns become particularly prominent in one especially promising use case of MR: remote collaboration between multiple users. In such scenarios, users may choose to share anything from virtual objects to entire representations of their local environments. Given that MR headsets are likely to be used in private spaces such as bedrooms, living rooms, and offices, it is crucial that users have the ability to control which parts of their surroundings they wish to share and which should remain private.

Currently, there are few functional solutions that offer users a comprehensive set of privacy tools in MR. Some approaches involve simplistic regional whitelisting or blacklisting via mechanisms like a “magic rope” (which involves cordoning off areas with a rope), while others use blurred regions that dynamically adjust based on user movement. However, these solutions are typically imprecise and heavy-handed, either removing too much or being visually intrusive. Moreover, most of these approaches either obscure or erase private regions without attempting to fill them with plausible replacement content, resulting in a disjointed viewing experience.

This challenge is addressed by the field of Diminished Reality (DR), described as the process of removing real-world objects from view and replacing them with background content to simulate seamless removal. Two main approaches exist: observation-based DR and inpainting-based DR. Observation-based methods rely on either multiple camera perspectives to reconstruct live 3D models of the environment, or on pre-scanned, detailed models of the space. However, such approaches are impractical for typical users due to cost and complexity. Moreover, even with a detailed 3D scan, the system might still lack the visual data needed to reconstruct what lies behind a specific object (e.g., a painting on a wall), rendering the method ineffective.

In contrast, inpainting-based DR offers a more robust and accessible solution for everyday users. With recent advances enabling real-time digital inpainting, a single secondary perspective can be used to process the scene while selectively removing specified private objects without the need for complex room models. A variety of inpainting techniques have been explored over the years, including traditional pattern-matching methods and, more recently, a range of deep learning approaches. These include generative adversarial networks (GANs), autoencoders, recurrent neural networks (RNNs), convolutional neural networks (CNNs), and most recently, transformer-based models.

In this thesis, I propose a novel privacy-control system that leverages real-time digital inpainting to allow users to selectively hide objects from their environment. Once identified, these objects are removed from the scene as viewed from a secondary perspective. Semantic segmentation and object recognition techniques are used to accurately delineate the boundaries of the objects to be removed, while a modified, performance-optimized transformer-based neural network performs real-time inpainting to fill the hidden regions with visually consistent content. This solution is robust, precise, and visually consistent, enabling the seamless integration of privacy tools into mixed reality applications for everyday users.
\chapter{Literature review\label{cha:litreview}}

An initial exploration of Extended Reality (XR) privacy systems was conducted to gain an understanding of the technical landscape and recognise areas that required further investigation.

\section{Privacy Systems and Their Necessity}
An abundance of research has been conducted on system-level considerations of user privacy in XR, investigating exactly which data points can be shared with applications, how the data is processed and how data is stored \cite{deguzman_2019_security}. As the technology has developed, researchers have recognised the risks of malicious AR apps recording sensitive information from users' surroundings \cite{jana2013finegrained} and have thus developed fine-grained permission systems to only share specific required data with applications. Studies evaluating perceptions of VR technology have also recognised that the public is generally concerned about the use and privacy of their spatial data \cite{Hornsey2024May}. Furthermore, some research has been conducted into the privacy concerns of multi-user shared-space MR meetings. Despite this, very few solutions exist that provide users with a set of tools to control which real-world items they wish to bring with them into the shared space. It is this tangible, user-controlled side of privacy that this paper's solution resides in.

\subsection{Concerns for Privacy in Shared-space MR}
The topic of privacy tools for mixed reality systems has been touched on for many years, initially in a hypothetical manner as the technology was being developed, and now in a more grounded fashion as the devices have matured and real applications can be tested on real users.

A key recent paper by \cite{lebeck2018security} attempts to provide a foundation for understanding the various challenges involved in the creation of mixed reality experiences from real user feedback. The study highlights a clear user need for privacy control tools in multi-user augmented reality environments. Participants expressed varying preferences for sharing content, with some favouring private experiences and others preferring default shared settings. Importantly, users suggested granular, object-wise sharing mechanisms such as “tap-to-share” interactions, designated shared zones, and visual indicators for content visibility. These suggestions were put forth by participants without being prompted, demonstrating an overt demand for clear and flexible access control features in AR systems. Through evaluation of these studies we can see that such a system must provide precise object-level privacy controls for users, which is what this paper will attempt to produce.

\subsection{Existing Multi-User Privacy Control Systems}
Various papers have outlined a set of guidelines for the design of privacy systems, such as the one by \cite{reilly2014secspace}. In this paper, the authors produced a loose set of criteria to guide future system design:
\begin{itemize}
  \item Use privacy mechanisms that are appropriate to the physical and virtual worlds
  \item Visually represent the current policies in both worlds
  \item Build on social norms when negotiating privacy mechanisms between the worlds
  \item Enforce privacy mechanisms based on context
  \item Provide simple authentication and permission controls
\end{itemize}
The resulting toolkit presented by the authors targets collaboration-specific scenarios such as shared whiteboards. By leveraging social norms and proxemics, the sharing/hiding of audio and whiteboard elements is restricted by physical distance to allow for rudimentary privacy controls in these spaces. This system does not specifically handle fine-grained object control for users deciding which real-world items they wish to bring into the shared space meeting, but is more directed towards shared virtual objects. Regardless, aspects of their system's design can be draw from.

In a recent paper by \cite{Ruth2019}, the authors analyse representative AR application scenarios to identify potential risks, producing design goals that ensure both security and functionality in multi-user settings. They introduce ShareAR, a prototype content-sharing control module implemented for the Microsoft HoloLens, which enforces access control policies and manages purely virtual content visibility among users. This work lays foundational principles for designing secure and private multi-user AR interactions, highlighting the importance of integrating security considerations into the core architecture of AR platforms. The module provides future researchers with a foundation for potential privacy control systems, however it does not explore the tangible functional side of completely removing real-world objects.

Various elementary interface designs have been explored that provide users with object-wise or regional sharing controls. An early key paper from \cite{butz1998vampire} explores what an interaction-based framework could look like for solving the privacy problem. They put forward two methods for visualising the privacy status of virtual items: a vampire mirror which reflects public items only, and a publicity/privacy lamp which the user can place over items to show/hide them. These systems leverage affordances to provide users with an intuitive interaction experience; an essential component in successful human-centred design.

Other interaction-based solutions include the design of the "magic rope" system by \cite{Qian2022Jun} which allows users to cordon off areas they wish to leave hidden with a flexible rope. They can specify public and private regions which can be exposed or excluded/blurred. This interaction system builds on existing social norms and affordances, resulting in a simple, accessible, and easily-understood interface. On the other hand, the system is relatively heavy-handed in nature and does not allow for fine-grained control, cutting out large regions of the users' surroundings. This does not align with all user preferences as found in the study mentioned earlier \cite{lebeck2018security}, as many users prefer object-level control and a more open, communal sharing experiences.

One solution that removes individual objects in a scene more specifically is from \cite{tabet_2024_blur}. In this paper, the author proposes a blurring-based object removal system. The blurred region dynamically grows and shrinks with user movement to ensure that the object remains obscured. However, the scaling of blurred regions is extremely aggressive and inevitably intrudes on the viewing experience. Furthermore, it has been found that various blurring methods are not entirely secure, as explored in a paper by \cite{hill_2016_on}. The authors concluded that many traditional blurring techniques are not viable approaches for content redaction as they can be easily recovered.

We can therefore determine from the existing literature that privacy systems are an integral part of multi-user mixed reality experiences. Few solutions specifically provide users with the ability to control what real-world items they wish to bring with them into a shared MR environment despite clear user need. Furthermore, existing solutions are heavy-handed, visually obtrusive, inconsistent and insecure, highlighting the need for a fine-grained, object-wise privacy tool that can remove individual private items. In order to meet these needs for multi-user MR applications, a new approach to object removal is required that is visually consistent, real-time and precise, ensuring that users can share their private spaces without compromising the privacy of their personal spaces. This is where the field of diminished reality can be explored.

\vspace{10mm}
\section{Diminished Reality}

Diminished Reality focuses on removing or concealing elements of the physical world so that users perceive a coherent, hole-free scene after those elements are gone. Two high-level strategies have dominated the field: observation-based techniques which reconstruct hidden content from additional viewpoints or an existing 3D scan of the space, and digital inpainting techniques which hallucinate the missing pixels directly. Considering the expense and additional calibration/setup/scanning that many observation-based approaches takes, inpainting-based techniques will be the primary direction of research, as this ensures that the system is both accessible and convenient for consumers.

\subsection{Digital Inpainting with Neural Networks}
With the growth of deep learning based approaches to various visual problems, neural networks began to be used in digital inpainting for their impressive capabilities in complex pattern recognition and scenic understanding. The feasibility of convolutional neural networks for inpainting was demonstrated by \cite{pathak_2016_context}, showing that these networks could recognise the structural elements of images and generate consistent results. A subsequent comprehensive review by \cite{Zhang2024Dec} outlined the characteristics and features of each neural network design:
\begin{itemize}
  \item \textbf{CNN-based encoders–decoders} capture local texture while preserving sharp edges.
  \item \textbf{RNN-based approaches} exploit temporal dependencies in video sequences.
  \item \textbf{GAN-based models} add an adversarial discriminator that encourages photo-realistic detail and have become the de-facto standard for complex textures.
  \item \textbf{Hybrid architectures} combine autoencoders, GANs, and recurrent attention to balance global semantics with fine texture.
\end{itemize}

Looking at more recent advancements, an inpainting model named E$^{2}$FGVI couples optical flow completion, feature propagation and content hallucination in a single trainable pipeline, pushing very high quality results but heavily reducing frame rates to \(<5\) fps due to the inefficient optical flow component \cite{li2022e2fgvi}.

One approach by \cite{kato2022online} combines both observation and inpainting-based DR into an incredibly capable system. It gradually builds a 3D reconstruction of the scene as the camera moves, shifting from pixel hallucination inpainting to an observation-based approach as more of the room is revealed. Despite the benefits of leveraging both methods, the complexity of the system denies any possibility of real-time application.

Despite their increase in perceptual quality, we can clearly see that many inpainting systems are far too bulky for real-time MR applications.

\subsection{Real-time Digital Inpainting}

Breakthroughs in efficiency gradually close the performance gap, as a subset of researchers specifically target real-time capable systems.

A prominent paper attempting real-time inpainting by \cite{herling2010advanced} put forward a patch-based system achieving \(\sim\)30 fps at a resolution of \(480\times360\), proving that interactive DR was feasible on standard hardware. However this rudimentary system would only provide acceptable results for small masks, simplistic planar scenes and mild camera motions.

\cite{Kobayashi2024Mar} incorporated channel attention into a Globally and Locally Consistent Image Completion model, trained with the aid of a GAN. The resulting pipeline could sustain decent frame rates \(\sim\)10 fps, however this could not be fully considered real-time for this application.

A key transformer-based model by \cite{Liu_2021_DSTT} named the \emph{Decoupled Spatial–Temporal Transformer} (DSTT) breaks attention into separate spatial and temporal streams.  This design cuts quadratic attention cost to linear, enabling \(720\text{p}\) inpainting at near real-time frame rates. DSTT preserves motion coherence better than earlier CNN-only baselines while remaining lightweight enough for edge-GPUs. This model was recently improved by \cite{thiry2024memory} who incorporate model 'memory' to improve temporal consistency by reusing past tensors. By 'remembering' what the scene looked like in previous frames, if the masked (hidden) region moves over an area that was previously viewed by the model, it can more accurately recreate the redacted region from memory. It is worth noting that these models were not designed for deployment, running PyTorch model code that did not take advantage of TensorRT optimisations.

Overall, DSTT currently offers the best trade-off between quality and speed for accessible hardware, making it a popular foundation for MR-centric prototypes.

\section{Object Detection and Tracking}

Fine-grained privacy control requires not only plausible background synthesis but also accurate, low-latency selection and segmentation of the objects to be hidden. User studies by \cite{cheng_2022_towards} show that participants prefer per-object toggles over region-wide redaction, citing a sense of agency and lower cognitive load. Building on that insight, \cite{tabet_2024_utilityprivacy} combine anchor-based tracking with semantic segmentation to produce more accurate object boundaries, while maintaining \(>\)15 fps. Their \emph{utility-privacy} framework highlights an important system-level balance: aggressive masking harms task performance, whereas precise boundaries maximise both frame rate and scene utility.

For the actual object detection and segmentation itself, various models have been developed, including R-CNN (Region-Based Convolutional Neural Network), introduced by \cite{girshick_2016_rcnn}. R-CNN was a pioneering object detection model that combined region proposal algorithms with deep learning. It first generates potential object regions using a selective search, then classifies each region using a CNN. Although it achieved high accuracy, R-CNN was computationally expensive and slow due to its multistage pipeline and repeated CNN evaluations for each region.

YOLO (You Only Look Once) by \cite{redmon2016yolo}, on the other hand, reformulated object detection as a single regression problem, predicting bounding boxes and class probabilities directly from full images in one forward pass through the network. This design allows YOLO to operate in real time, offering significantly faster inference speeds compared to R-CNN. Recent advancements with YOLO v11/12 have made improvements to the speed, size, and accuracy of the model \cite{tian2024yolov12} which make it a perfect option for a real-time DR
application.

\bigskip
\noindent
\section{Summary}
The evolution from multi-camera observation systems to transformer-based real-time inpainting has brought DR to the cusp of consumer-grade mixed reality headsets. Yet no existing work offers an end-to-end toolchain that couples state-of-the-art object segmentation, fast transformer inpainting and an MR user interface for on-demand privacy management. The remainder of this thesis addresses that gap.
\chapter{Methods\label{cha:methods}}

In order to produce a privacy system capable of real-time object redaction, various components need to be implemented:
\begin{itemize}
    \item The scene needs to be captured.
    \item Objects need to be recognised and presented to the viewer for selection.
    \item Real-time diminished reality needs to be enabled through segmentation and digital inpainting.
    \item A sophisticated system for handling these complex processes and interactions needs to be designed.
\end{itemize}

The project's scope was initially defined through clear success criteria outlined below. This approach ensured that the final outcome would function effectively as a privacy system while also preventing time from being spent on unnecessary components.

\section{Success Criteria}

By drawing from the relevant literature, the system was designed to meet the following criteria for a functional mixed reality (MR) privacy tool:

\begin{enumerate}
    \item Real-time performance (\(\geq\)20 fps)
    \item Accurate per-object redaction capability
    \item Plausible scene reconstruction using local visual context
    \item Robust handling of dynamic camera perspectives
    \item Low-effort interaction (e.g., tap-to-hide)
    \item Clear visual feedback of current privacy settings
\end{enumerate}

Real-time performance is essential for any headset-based application, as low frame rates can cause user discomfort and disorientation \cite{zhangsickness2020}. As discussed in the literature review, users in multi-user MR environments typically prefer per-object redaction capabilities to preserve agency and reduce cognitive load. Plausible scene reconstruction is critical to avoid visually intrusive or distracting results. Given that users may reorient their headsets or camera setups during operation, the system must be capable of handling camera movement. From a human-computer interaction (HCI) standpoint, interaction mechanisms should leverage simple, familiar affordances to support intuitive use. Finally, current privacy policies should be visually represented within the environment to support user awareness and trust.

\section{Approach}

The system operates using a single depth camera positioned to capture the primary user's environment. This camera generates a live stream of RGB and depth data, forming a dynamic 3D representation of the scene which can be shared with a remote user as a point cloud.

To enable privacy control, object detection is first performed on the live camera feed. Detected objects are transmitted to the primary user's MR headset as bounding boxes (a virtual 3D box encompassing the object, with real-world positional coordinates), overlaid within their spatial view. If the user wishes to redact a particular object, they can simply select it by pointing and clicking on its bounding box.

Once an object is selected for redaction, the system segments the object in the capture stream and performs digital inpainting to plausibly fill the missing region. The modified image and corresponding 3D data are then transmitted to the remote user, effectively removing the private object from their view of the scene.

\newpage
\section{System Overview}

\begin{figure}[htbp]
    \centering
    \includegraphics[width=\linewidth]{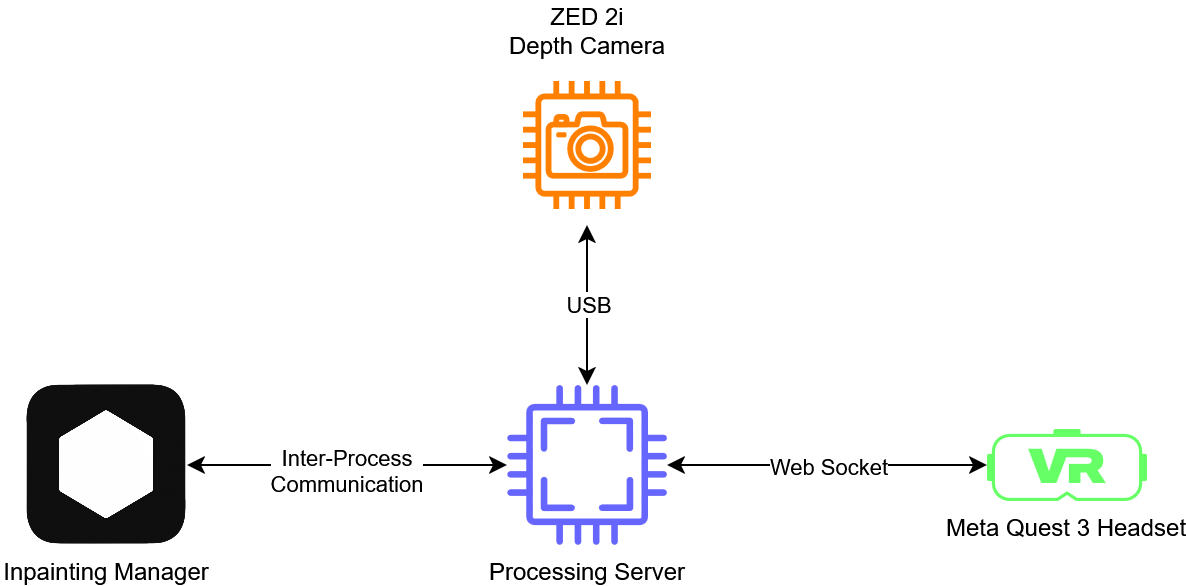}
    \caption{Simple System Overview}
    \label{fig:simple-system}
\end{figure}

The system is composed of three main modules:
\begin{itemize}
  \item \textbf{Processing Server} (C++): Core logic, object detection, ZED camera handling, redaction control.
  \item \textbf{Inpainting Manager} (C++): Dedicated real-time inpainting engine using GPU-accelerated transformer models.
  \item \textbf{Headset Client} (Unity/C\#): Receives object bounding boxes and sends user selections, displays UI overlays.
\end{itemize}
Each is described below.

\subsection{Processing Server}
\subsubsection*{Purpose}
The processing server acts as the central C++ program which interacts with all of the various components. It handles which objects are to be removed, manages the object detection/segmentation pipeline, and visualises the results. It actively manages the ZED API which controls camera initialisation, frame/depth retrieval and object ingestion. The ZED API exposes various essential methods for object tracking and depth measurement. These are used to convert found objects from the 2D image into a 3D coordinate system.

\subsubsection*{Object Recognition/Segmentation}
YOLO v11 was selected for object recognition and image segmentation. It is the most recent YOLO release that exports a segmentation-specific model, and it demonstrates capable and accurate object recognition on the COCO dataset. The following shows the measured accuracy and speeds of each size model for YOLO v11:

\begin{figure}[h!]
\centering
\begin{tikzpicture}
\begin{axis}[
    width=0.8\textwidth,
    xlabel={YOLO Version},
    ylabel={mAP50-95},
    ymajorgrids=true,
    xtick=data,
    xticklabels={YOLO11n, YOLO11s, YOLO11m, YOLO11l, YOLO11x},
    ymin=0.65, ymax=0.82,
    yticklabel style={text width=2em,align=right},
    legend style={at={(0.5,-0.2)},anchor=north,legend columns=2},
    axis y line*=left,
    axis x line=bottom,
    enlarge x limits=0.1
]
\addplot+[blue,mark=*] coordinates {(0,0.668) (1,0.742) (2,0.795) (3,0.794) (4,0.784)};
\addlegendentry{mAP50-95}

\end{axis}

\begin{axis}[
    width=0.8\textwidth,
    ylabel={Total Time (ms)},
    ymin=0, ymax=8,
    axis y line*=right,
    axis x line=none,
    ymajorgrids=true
]
\addplot+[red,mark=square*] coordinates {(0,2.2) (1,2.6) (2,4.0) (3,4.6) (4,7.4)};
\addlegendentry{Total Time}
\end{axis}
\end{tikzpicture}
\caption{Comparison of YOLO11 versions by accuracy and inference time.}
\end{figure}
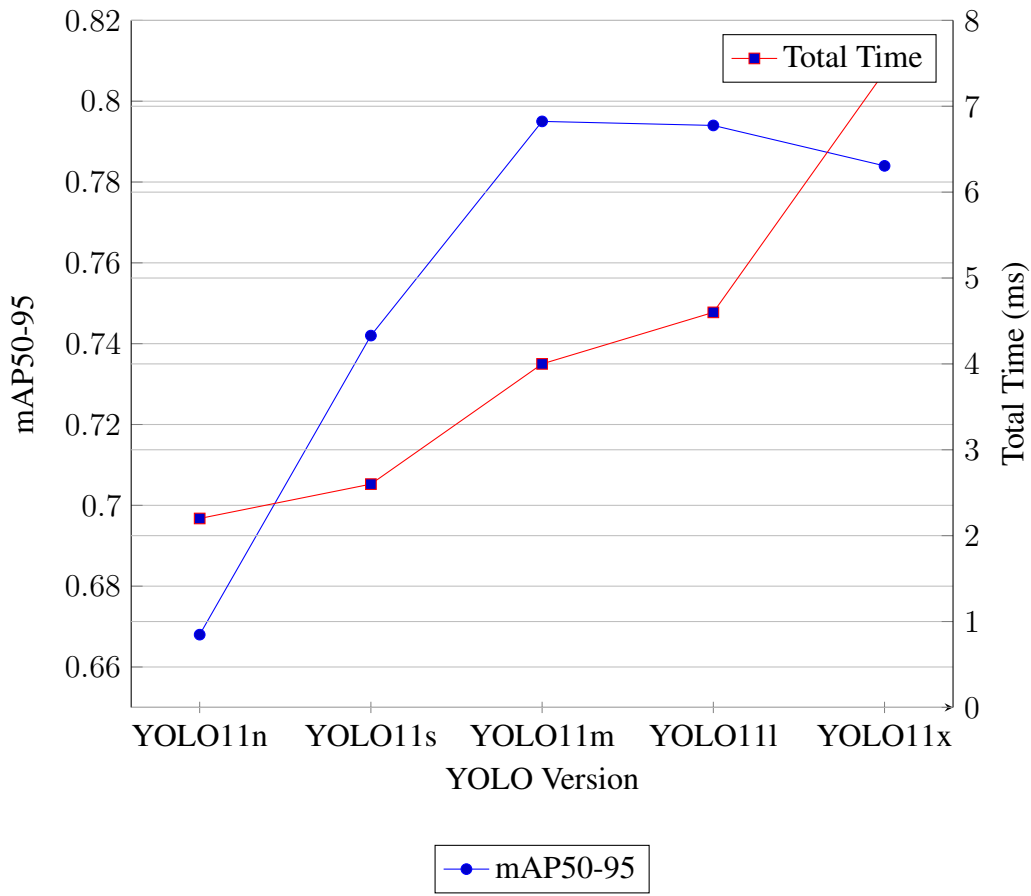

The medium-sized model was utilised for its balance between speed, size and accuracy.
The raw frame is sent to the model which returns masks and bounding boxes for precise selection and object-wise redaction, meeting criteria (2).

\subsubsection*{I/O}
Communication with the headset is done through web sockets. By maintaining a listening thread and exposing a communication port, the system is able to send bounding box JSON information to the headset each frame, and can receive essential object selection information in return. By placing the listener on a separate thread, inpainting can continue to take place while a change queue stores any messages from the headset user, ensuring that no object selections are missed.

Due to GPU contention and instability when running both object detection and inpainting networks within the same process, the inpainting pipeline was offloaded to a separate service. (See Appendix~\ref{appendix:multi-model} for a more detailed breakdown of the issue)

Communication with the inpainting manager service is done through Inter-Process Communication (IPC). A reserved portion of device memory is shared between the two processes, which is used for passing back and forth the raw/inpainted frames. Signals are used to notify the other process when the writing process is complete and reading can begin.

\subsubsection*{Object Control}
The server manages a list of private objects that the user wishes to be removed. Each frame, it takes the relevant object masks from the segmentation algorithm and cuts the objects out of the frame leaving empty regions. This item-redacted frame is then passed to the inpainting model for pixel generation.

\subsubsection*{Point Cloud Visualisation}
Resulting inpainted point clouds are visualised with OpenGL, a platform for rendering graphics. The initial point cloud is modified by transplanting the RGB values from the privatised inpainted frame onto the depth measurements, creating a rendered 3D representation of the room with items effectively removed.

\subsection{Inpainting Manager}
The inpainting manager hosts the inpainting model itself. 
Systematic exploration of various inpainting technologies was carried out, and thus various solutions were tried and disposed of in the process of selecting an ideal model candidate. This process of testing can be viewed in Appendix ~\ref{appendix:inpaint-choices}. As a result, the DSTT model designed by \cite{Liu_2021_DSTT} was chosen for its lightweight footprint, simplistic yet robust hybrid CNN/transformer architecture, and balance between speed and accuracy. It also boasts temporal improvements incorporated by \cite{thiry2024memory} which utilise 'memory' tensors of previous frames for a more consistent and accurate result.  The model accepts a frame with masked regions set to a value of 0, and two existing memory tensors from previous outputs.
It then outputs a completed frame; redacted regions filled with hallucinated background content, alongside a new memory tensor. The content creates plausible background geometry meeting requirement (2).

\subsection{Headset Client}
\subsubsection*{Perspective Alignment}
In order to view the 3D bounding boxes generated by the system, the ZED camera perspective and the headset perspective need to be aligned within 3D space. After attempting to build a solution with the use of positional markers, a far more precise and simplistic approach was chosen. The user places a semi-transparent virtual ZED camera over the real camera, encompassing it and aligning it visually. We then compute a rigid body transformation between the two coordinate systems, composed of a rotation matrix \( \mathbf{R}_{\text{zed}} \in \mathbb{R}^{3 \times 3} \) and a translation vector \( \mathbf{t}_{\text{zed}} \in \mathbb{R}^3 \). These map 3D points from the ZED camera's world space into the headset's world space.

\vspace{10mm}
Let:
\begin{itemize}
    \item \( \mathbf{p}_{\text{bbox-zed}} \in \mathbb{R}^3 \): a point in the ZED camera's world coordinates (e.g., a bounding box center).
    \item \( \mathbf{p}_{\text{bbox-head}} \in \mathbb{R}^3 \): the corresponding point in the headset's world coordinates.
\end{itemize}

\textbf{Calibration Step}

To compute the relative pose of the ZED camera with respect to the headset, we find the positional and rotational transformation of the virtual ZED box relative to the headset's neutral position.

\[
\mathbf{p}_{\text{zed-in-head}} = \mathbf{R}_{\text{head}}^{-1} (\mathbf{p}_{\text{zedbox}} - \mathbf{t}_{\text{head}})
\]
\[
\mathbf{R}_{\text{zed-in-head}} = \mathbf{R}_{\text{head}}^{-1} \mathbf{R}_{\text{zedbox}}
\]

\newpage
We then take the current position of the headset and calculate the pose of the ZED camera within the Unity world space:

\[
\mathbf{t}_{\text{zed}} = \mathbf{R}_{\text{head}} \, \mathbf{p}_{\text{zed-in-head}} + \mathbf{t}_{\text{head}}
\]
\[
\mathbf{R}_{\text{zed}} = \mathbf{R}_{\text{head}} \, \mathbf{R}_{\text{zed-in-head}}
\]

\textbf{Transforming Detected Object Positions}

Each detected object's position \( \mathbf{p}_{\text{bbox-zed}} \) in the ZED camera coordinate system is transformed into Unity world coordinates using the relative pose data:

\[
\mathbf{p}_{\text{bbox-head}} = \mathbf{R}_{\text{zed}} \, \mathbf{p}_{\text{bbox-zed}} + \mathbf{t}_{\text{zed}}
\]

This transformation ensures that bounding boxes detected by the ZED camera are correctly located and rendered in the headset’s spatial frame, meeting requirement (4).

\subsubsection*{Object Bounding Box Overlay}
To visualise the items for selection, virtual bounding boxes are placed around each object in the scene. In order to reflect the current privacy setting for that object as per requirement (6), the colour of the box is used to represent its state. Social norms are leveraged to ensure that users easily understand the state of the privacy settings. Green is used to represent public items, and red is used to represent private items that are being hidden.

A custom shader was designed which emphasises the edges of the bounding box, also including a transparent centre to view the contents. This ensures that the contents can still be easily viewed by the user while still clearly representing its privacy state.
The boxes are designed to fit naturally into the environment and reduce visual clutter. This is achieved through depth-based object occlusion that ensures that the bounding boxes become hidden when obscured by real-world content.
\begin{figure}[htbp]
    \centering
    \includegraphics[width=8cm]{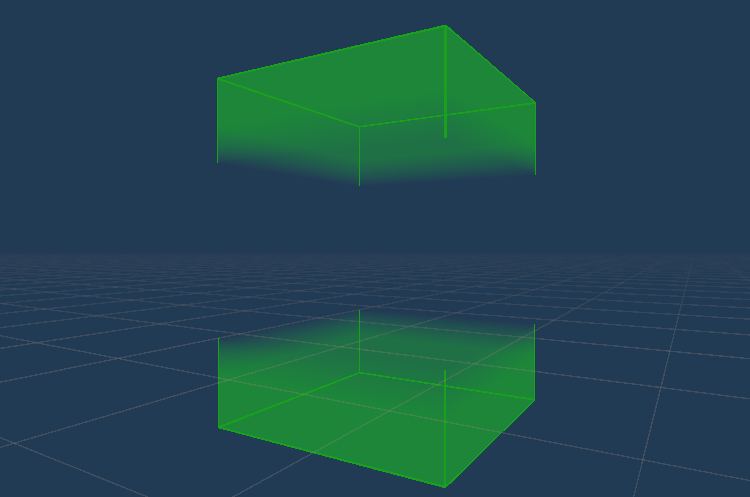}
    \caption{Bounding Box Prefab}
    \label{fig:bounding-box}
\end{figure}

\newpage
\subsubsection*{Interaction Design}
Simple point-and-click toggling of item privacy settings offers simple and intuitive user interactions with the system. Changes in privacy settings are immediately reflected visually by a change of colour, and the preference is sent to the system to begin object redaction. The point-and-click approach matches user expectations and preferences outlined in requirement (5).

Having outlined each core component, the following section presents the full processing pipeline that integrates them.

\newpage
\section{System Pipeline}

\begin{figure}[htbp]
    \centering
    \includegraphics[width=\linewidth]{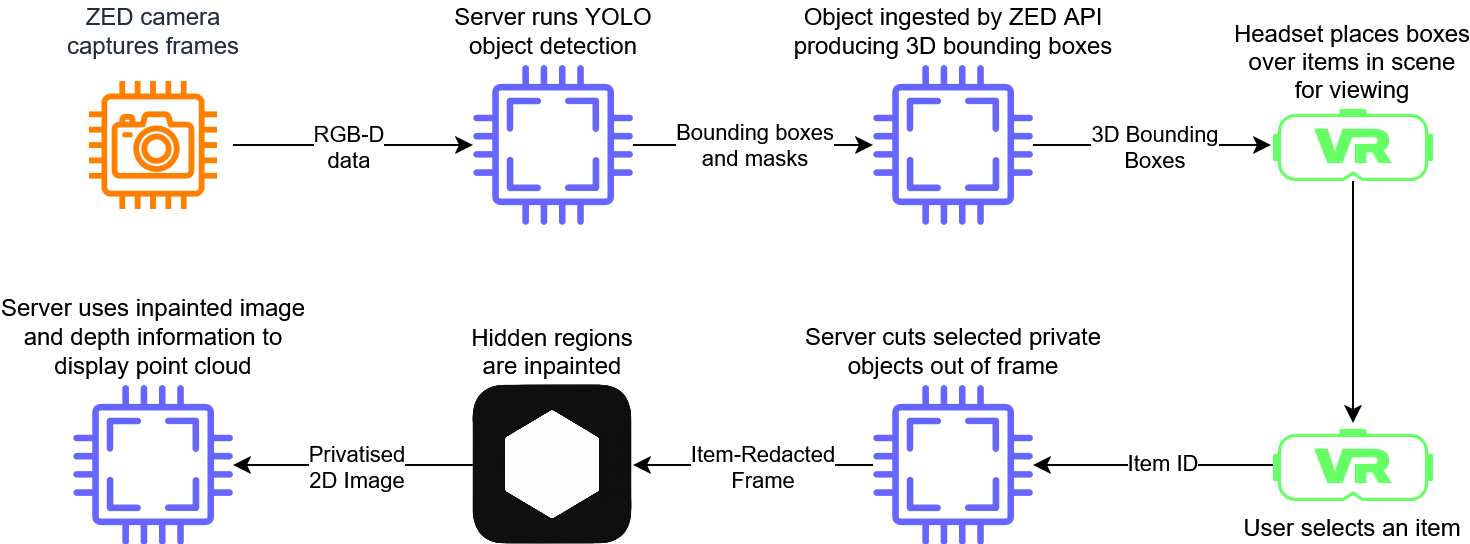}
    \caption{System Flow Diagram}
    \label{fig:flow}
\end{figure}

\begin{figure}[htbp]
    \centering
    \includegraphics[width=15cm]{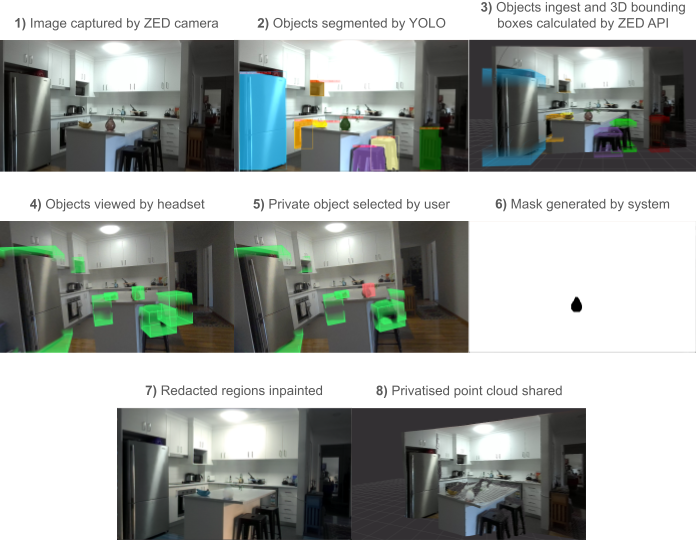}
    \caption{System Flow Demo}
    \label{fig:flow-demo}
\end{figure}

\autoref{fig:flow} and \autoref{fig:flow-demo} show an overview of the system pipeline, including a visual demonstration of the pipeline processing an example setting.


\section{Interaction Flow}
The following describes the series of interactions a user will have with the system to privatise certain objects.

\subsection*{Startup and Perspective Alignment:} 
\begin{figure}[htbp]
    \centering
    \includegraphics[width=15cm]{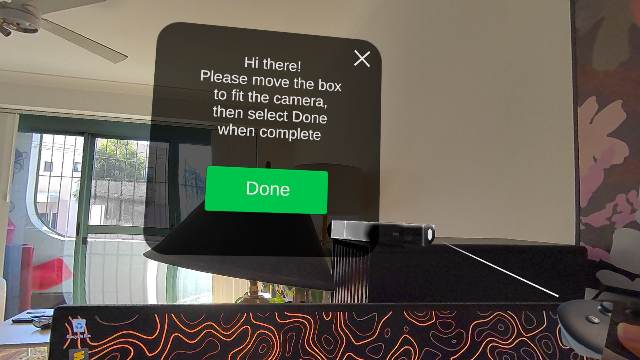}
    \caption{Perspective Alignment}
    \label{fig:alignment}
\end{figure}

Upon starting the application on the headset, the user is greeted with a welcome banner and a small grabbable ZED box. In order to align the ZED camera and headset camera perspectives, the user must pick up and place the virtual ZED bounding box over the real ZED camera. This is done with simple click-and-drag mechanics. Once positioned, the 'Done' button on the banner can be pointed at and clicked to confirm the position. This can be seen in \autoref{fig:alignment}.

\subsection*{Private Object Selection:} From that point, objects discovered by the main system will then become visible to the headset user. Green bounding boxes encompassing the various potential items will be displayed. To make an item private from external view, the user simply points at a box and clicks (presses the remote trigger). This will turn the box from green to red, indicating its change in state. This action will be sent to the system which then begins inpainting that particular region from the secondary perspective. To make the item public again, the same point-and-click action can be completed to change the colour back to green, and the system will no longer hide that particular object.
\subsection*{Recalibrating ZED Position:} In the event that the objects are not aligned correctly, the user can select the virtual ZED camera box and move it to a more correct position. The welcome dialogue box will re-appear, allowing the user to press confirm again when repositioned.

\section{Performance Optimisations}

Considering this pipeline needs to be run at a minimum of 20 fps as per requirement (1), performance optimisation has been an extremely important focus in the implementation of the system.

\subsection{Upscaling}
Both the segmentation and inpainting models are run at a lower resolution of 640x360, and their outputs are then scaled by a factor of 2 to 1280x720p. This allows the models to run more efficiently at the cost of slight pixelation in inpainted regions, and slightly jagged object outlines. The impact has been found to be minimal. 

\subsection{Inpainting Model}
One of the most intensive components of the system is the inpainting model itself, and thus considerable effort has been made to improve model performance.

The original DSTT model is built as a PyTorch model without any particular performance optimisations. It is mainly designed for experimentation and research, and is not prepared for ONNX/TensorRT export. Memory modifications to the model made by \cite{thiry2024memory} did not make this improvement, instead focussing on adding memory tensors to the input/output of the model to improve accuracy. Although the model's architecture itself is designed with real-time performance in mind, inference speeds could be improved through the use of TensorRT optimisations. It has been found that in most cases, Nvidia's TensorRT platform can increase the performance of various neural networks through techniques such as layer fusion, quantisation, dynamic tensor memory management, kernel auto tuning and precision calibration \cite{tensor_gopik_2024}. By modifying the model for export to TensorRT, performance benefits can be expected.

The architecture of the model included various non-linear and 'pythonic' characteristics that made export extremely difficult. In order to prepare the model for export, the following modifications were made that converted the fairly dynamic model to a more static one.
\begin{itemize} \label{DSTT-mods}
    \item \textbf{Fixed input resolution:} By fixing the resolution to 640x360, the model does not have to dynamically adjust to input size.

    \item \textbf{Minimal memory slots:} Instead of accommodating a varying number of memory slots, the model was adjusted to handle a fixed number of memory tensors. Testing was undertaken to find a balance between performance and accuracy in the number of memory slots. The number of slots was modified, and the resulting FPS was measured:
        \begin{table}[htbp]
            \centering
            \begin{tabular}{|c|c|c|c|}
                \hline
                \textbf{Memory Slots (m)} & \textbf{Frame Time (ms)} & \textbf{Raw FPS} & \textbf{Real FPS} \\
                \hline
                2 & 14.7 & 68 & 30 \\
                3 & 18.6 & 54 & 20 \\
                4 & 23.0 & 43 & 17 \\
                \hline
            \end{tabular}
            \vspace{6pt}
            \caption{Performance metrics for different memory slot counts. Real FPS includes camera-to-tensor conversion and file access overheads.}
            \label{tab:memory-performance}
        \end{table}

    \noindent
    As a result, 2 memory slots were selected for use as this provided the model with adequate temporal accuracy whilst maintaining high frame rates.
    
    \item \textbf{Modified Vec2Patch Algorithm:} The original model utilised folding for reconstructing tensors post-transformation. Although beneficial for inpainting due to the summing of overlapping regions which captures relationships between vectors, this operation is not supported by TensorRT and thus a different approach was designed. A reduction convolution approach was used with a final crop to ensure the resulting vector was of the correct size.
    
    \item \textbf{Distinct Transformer Blocks:} The original model dynamically interleaved 2 different transformer blocks; a spatial transformer block and a temporal transformer block. This dynamic addition of layers does not work with TensorRT and thus a fixed alternative series of blocks was used instead. More explicit memory management was designed and the grouping of the frame into multiple blocks for the temporal transformer was more explicitly defined. Furthermore, the transformer block's attention mechanism was slightly modified, moving from a custom mechanism to one utilising PyTorch's \(scaled\_dot\_product\_attention\) for reliability and speed.
\end{itemize}

Once modified for export to ONNX and then to TensorRT, FP16 precision was utilised for its faster inference speeds and smaller memory footprint.

Due to modifications made to the internal architecture of the model, complete re-training was required.

\section{Model Training}

\subsection{Dataset selection}
The original DSTT model was trained on a mix of the YTVOS and DAVIS datasets. These collections contain a wide variety of different video environments, contents, and actions. Considering the application of this system is generally for indoor use, three alternative datasets were selected that mirror the potential environments that the system should be familiar with:
\begin{itemize}
    \item \textbf{GTA-V:} This dataset, introduced by \cite{richter2016playingdatagroundtruth}, is rendered within the Grand Theft Auto V video game and features a wide variety of scenes, including both moving and fixed-perspective shots of individuals performing actions in diverse environments. The camera distance closely resembles that expected in this application, and the inclusion of people aligns well with the intended deployment context. Additionally, the smooth camera motion and consistent image quality—due to the virtual nature of the recording—make it an ideal, low-complexity starting point for training.
    
    \item \textbf{MIT Indoor Scenes:} This dataset from \cite{quattoni2009indoor} was constructed by to address the challenge of categorising indoor scenes. Containing 15620 images under 67 different categories, this dataset is expansive, which would assist in exposing the model to a wide variety of spaces. Because the dataset only contains individual frames and not videos, each image must be duplicated to form a short 5-frame sequences with simulated camera movement. This ensures that the model learns to adapt to slight movements over time, improving its temporal accuracy.

    \item \textbf{SUN3D:} This dataset was created by \cite{xiao_sun3d_2013}, and is a large-scale RGB-D video database containing both camera pose information and object labels. Initially designed to train 3D space mapping algorithms, the videos capture the full 3D geometry of various 3D environments. This dataset provides an excellent addition for training as it will help expose the model to a myriad of potential settings of varying complexity at different camera angles.
\end{itemize}

\subsection{Mask generation}
To generate masks that redact part of the image for inpainting, a combination of two approaches were used. The first involves propagating a single stable mask across all frames in the sequence, and the second generates different masks each frame to simulate movement. This helps the model learn to use its provided frame memory, whilst also ensuring it can still inpaint purely based on surrounding content. 

Each mask shape is generated randomly each frame either by creating a rectangle or a more complex shape involving multiple random strokes. In either case, the calculated mask area is then diluted or eroded to ensure that the region covers between 5 and 30\% of the frame. This is done to expose the model to a variety of redacted object shapes.

\subsection{Training structure}
The training was completed in two stages to form a curriculum. Research by \cite{hacohen2019powercurriculumlearningtraining} has outlined the benefits of curriculum-based training for deep networks and image-based CNN's, demonstrating increased learning speeds and improved final performance on test data. The first stage combined the GTA-V and MIT Indoor datasets, and the second stage involved just the SUN3D dataset. This approach was selected so the model could easily establish a baseline from the more controlled, less complex and explicitly textured first two datasets, and then step up the difficulty to a more dynamic and less textured dataset to include more complexity. The ramping up from simple to complex data gives the model time to learn simple features initially, before training the temporal and memory-based aspects with complex camera movement and more difficult scene content. The first stage was relatively small, with a size of 3.8 GB, and the second stage was much larger with a size of 41 GB. This meant that far more epochs were used to work through stage 1 until results stabilised, and fewer epochs were used for stage 2.

\textbf{Loss:} L1 loss was used to evaluate the accuracy of both the masked region and the rest of the image. The total loss is weighted towards the masked region results to ensure that the model focusses on the redacted region primarily while learning. Loss figures are calculated on a per-pixel basis comparing the hallucinated pixel with the ground truth.

\textbf{Learning Rate:} A standard learning rate of 1e-4 was used for stage 1, and a slower learning rate of 5e-5 was selected for stage 2 to essentially 'fine-tune' the model weights as results were already fairly decent. The optimizer was also reset for stage 2 to reset the various gradients for the new dataset.

\textbf{Batch Size:} A batch size of 2 was selected. Although this seems small, each batch contained a sequence of 5 frames to train the memory component of the model. As a result, the resulting VRAM footprint was still quite high, maximising the effectiveness of each pass.

\begin{figure}[htbp]
\centering
\begin{tikzpicture}
\begin{axis}[
    xlabel={Epoch},
    ylabel={Training Loss},
    title={Training Loss Over Epochs},
    legend style={at={(0.5,-0.35)}, anchor=north, cells={align=left}},
    width=0.9\linewidth,
    height=6cm,
    grid=major
]
\addplot+[mark=*, thick] coordinates {
    (1,0.1444) (2,0.0979) (3,0.0888) (4,0.0837) (5,0.0773)
    (6,0.0741) (7,0.0718) (8,0.0703) (9,0.0687) (10,0.0673)
    (11,0.0658) (12,0.0646) (13,0.0638) (14,0.0629) (15,0.0622)
};
\addlegendentry{Stage 1}

\addplot+[mark=square*, thick, dashed] coordinates {
    (16,0.0444) (17,0.0425)
};
\addlegendentry{Stage 2}
\end{axis}
\end{tikzpicture}
\caption{Training loss per epoch across two training stages.}
\label{fig:training-loss}
\end{figure}
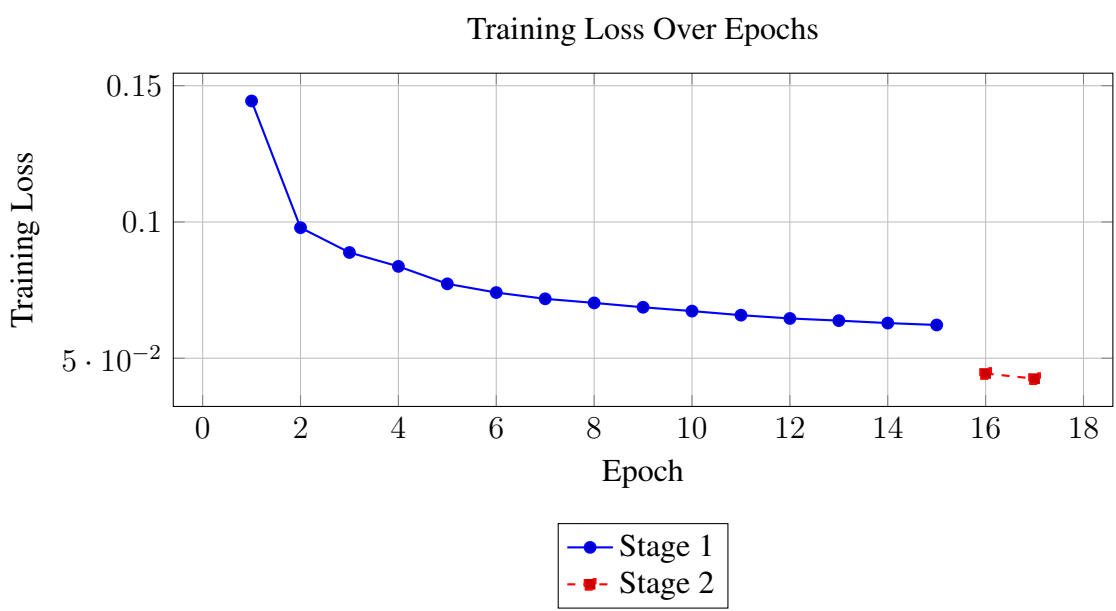

Figure~\ref{fig:training-loss} shows the training loss progression. Note that further training could be completed for improvements in accuracy. The training process was stopped at an acceptable point due to cost and time restrictions, however the loss was still trending downwards at the point of stoppage.

The resulting trained model was then exported as a TensorRT engine for inpainting within the system.

\section{Tools and Libraries Used}
The following tools, libraries, APIs and software were used in the development of the system.

\subsection{Software}
Unity was used to developed the headset application. Visual Studio was used for the development of the inpainting server and processing server applications in C++.
\subsection{APIs and Libraries}
The ZED SDK was utilised for its handling of spatial data and tracking of objects from frame to frame. OpenGL was used to visualise the point clouds. OpenCV was used for manipulating the visual data. The code for interacting with the YOLO model at the hardware level, and the OpenGL visualisation code was repurposed from the Stereolabs example codebase.

\subsection{Hardware}
The MR headset used for the system is a Meta Quest 3, however any headset with passthrough and Wi-Fi capabilities could be used. A ZED 2i depth camera is used as the main capturing device. A fairly capable desktop PC was utilised to run the majority of intensive processing, including the inpainting and segmentation models. The PC has the following specifications:
\begin{itemize}
    \item \textbf{GPU:} NVIDIA GeForce RTX 3070 with 8GB VRAM
    \item \textbf{CPU:} Ryzen 7 7700X 8-Core
    \item \textbf{RAM:} 32 GB
\end{itemize}

\section{Inpainting Limitations}
There are a variety of limitations to the current implementation which could be improved or modified with further work. Various trade-off's were encountered, and in order to achieve some goals, some sacrifices were made.

One such sacrifice was in the quality of inpainting results, for the purpose of achieving real-time throughput. Although other models achieve higher fidelity, they do not support real-time performance, necessitating quality-speed trade-offs. Thus the decision was made to slightly diminish inpainting quality to improve the speed of inference.
Furthermore, the memory functionality was heavily impacted by the necessary reduction of memory tensors. The original model had a more extensive memory of previously seen frames, which provided the model with maximum temporal information. Because of the optimisations made to the model for the purpose of speed, only 2 memory frames are provided as reference rather than the initial 5-10.

Despite these limitations, the system achieves a balance between perceptual quality and speed, making it suitable for real-time MR privacy control in typical indoor environments.

\section*{Source Code}
The full system implementation, including inpainting manager code, Unity scripts and the main processing server code is available on GitHub; see Appendix~\ref{appendix:sourcecode} for the repository links.
\chapter{Results\label{cha:results}}

The system successfully enables the selective redaction of private objects from remote perspectives in real-time MR. The following section describes how the criteria for success were met, and outlines the capabilities of the system in terms of its performance, quality, and accuracy.

\section{Requirements Evaluation}
\begin{enumerate}
    \item \textbf{Real-time performance:} Achieved \(\sim\)21 fps on RTX 3070 (goal \(\geq\) 20 fps).
    \item \textbf{Per-object redaction:} Fully implemented using bounding-box selection and YOLO detection.
    \item \textbf{Plausible scene reconstruction:} Semi-realistic inpainting achieved.
    \item \textbf{Dynamic camera handling:} Successfully manages a moving ZED camera. The perspectives remain aligned.
    \item \textbf{Low-effort interaction:} Simple and robust point-and-click interface implemented.
    \item \textbf{Visually represent current policies:} Current privacy settings visualised with minimally-obtrusive bounding boxes.
\end{enumerate}

The results indicate that all key requirements have been met to a satisfactory degree, validating the system’s core design. The degree of success can be further evaluated through the analysis of system performance and critique of visual results.

\subsection{System Performance}
\begin{figure}[htbp]
    \centering
    \includegraphics[width=\linewidth]{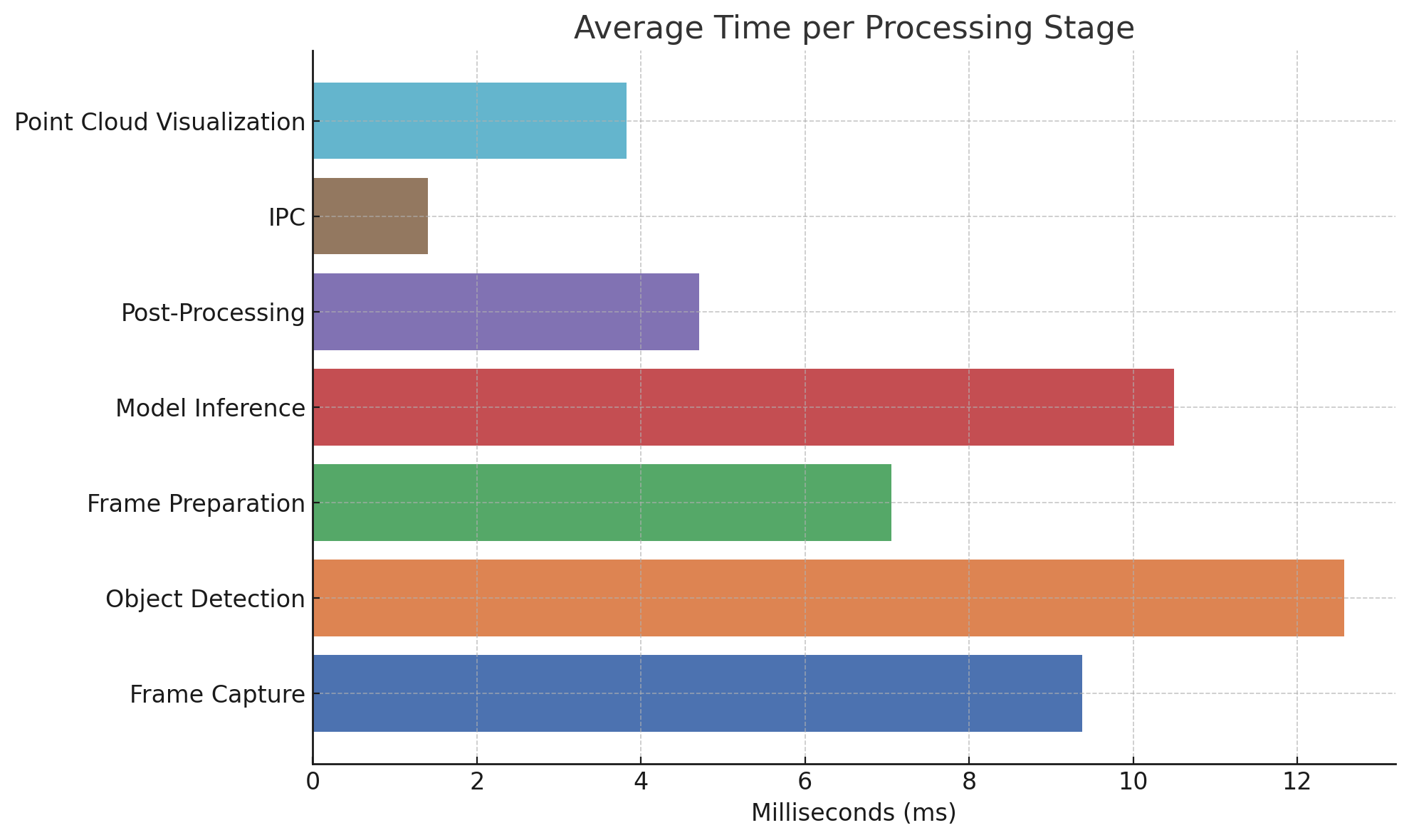}
    \caption{System Timing Breakdown}
    \label{fig:system-timing}
\end{figure}

To evaluate the performance of the system, average per-frame processing times were recorded for each major stage of the pipeline. These include frame capture, object detection, inpainting (broken down into preparation, model inference, post-processing, and IPC overhead), and point cloud visualisation. The timing measurements were collected across multiple frames on an RTX 3070 GPU and are summarised in Figure~\ref{fig:system-timing}.

As shown, the object detection and inpainting stages dominate total frame time, with inpainting alone accounting for over 20 ms due to CPU pre/post processing and GPU inference. IPC overhead and point cloud visualisation contribute smaller, but non-negligible delays. These results confirm that the system meets real-time constraints (\(\geq\)20 fps), while highlighting key performance bottlenecks.

The extremely fast inference speeds of the inpainting model represent a key performance success. Reducing inference times down to \(\sim\)10.5 ms plays an essential part in realising the real-time system goal. For comparison, the original DSTT model took \(\sim\)65 ms to generate a single frame at the same resolution, demonstrating an 84\% reduction in inference time with the various optimisations.

Beyond numerical timing results, it is important to evaluate the system from the perspective of user experience and visual fidelity.

\section{Visual Results}
Visual results were evaluated subjectively in multiple spaces containing a variety of different objects. Extensive multi-user testing and evaluation was not possible due to time constraints for the project.

\subsection{User Interface}
The bounding box visualisation is clear and moderately non-obtrusive. However, due to the inaccuracies and non-deterministic nature of YOLO object detection/segmentation, the bounding boxes can move about in a jittery manner and sometimes jump in and out of view. A more stable bounding box system would be less visually obtrusive. Furthermore, smaller objects do not 'fit' the bounding boxes well, resulting in a more solid box rather than a semi-transparent gradient for better visibility.

Despite minor visual issues, the interaction system is intuitive and simple, allowing for rapid selection of private items.

\subsection{Remote Perspective}
\subsubsection*{2D Render}
\begin{figure}[htbp]
    \centering
    \includegraphics[width=\linewidth]{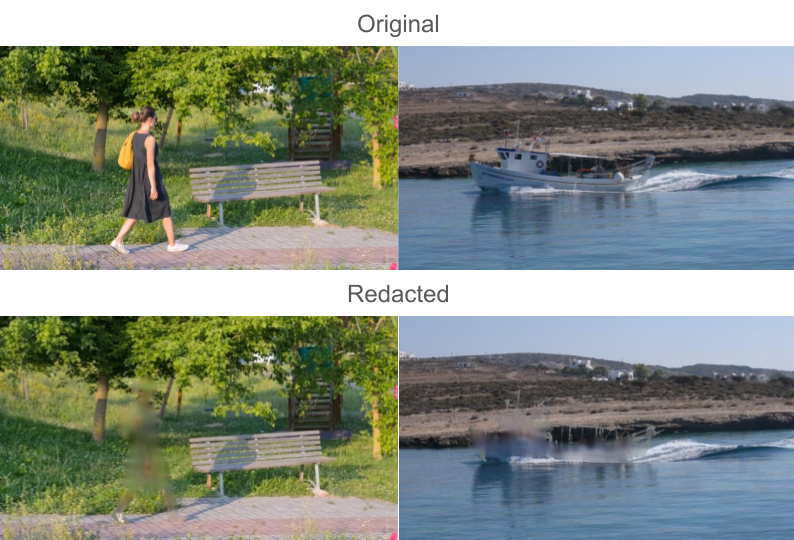}
    \caption{Inpainting results on DAVIS 2}
    \label{fig:inpaint-lucia-boat}
\end{figure}

Inpainting quality is moderate, producing visually plausible results in simple scenes. They produce somewhat blurred and imprecise results, but the model does well to recognise colours, scene geometry and features, and is capable of continuing planes and lines through inpainted regions. Simplistic environments such as single-colour backgrounds result in higher quality results, however complex backgrounds can often not be inpainted accurately as the content is more difficult to generate. \autoref{fig:inpaint-scoot} and \autoref{fig:inpaint-lucia-boat} demonstrate the performance of the model on various images from the DAVIS dataset \cite{davis2016}:

\begin{figure}[htbp]
    \centering
    \includegraphics[width=\linewidth]{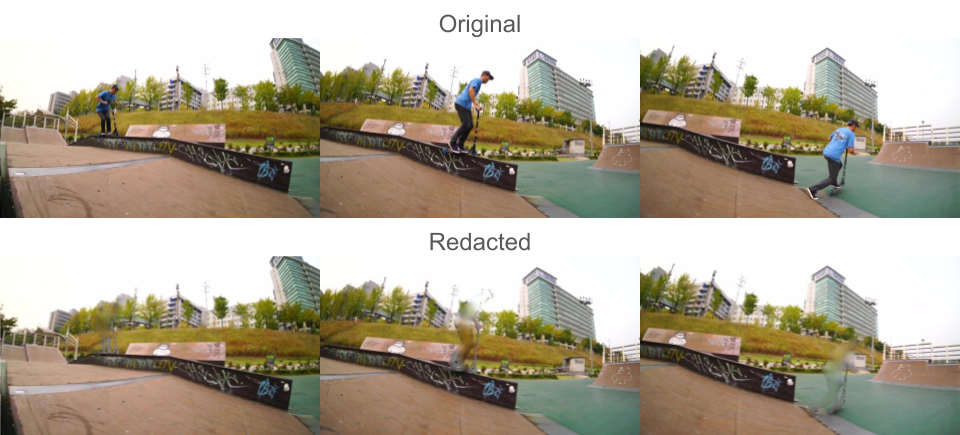}
    \caption{Inpainting results on DAVIS 1}
    \label{fig:inpaint-scoot}
\end{figure}

\begin{figure}[htbp]
    \centering
    \includegraphics[width=\linewidth]{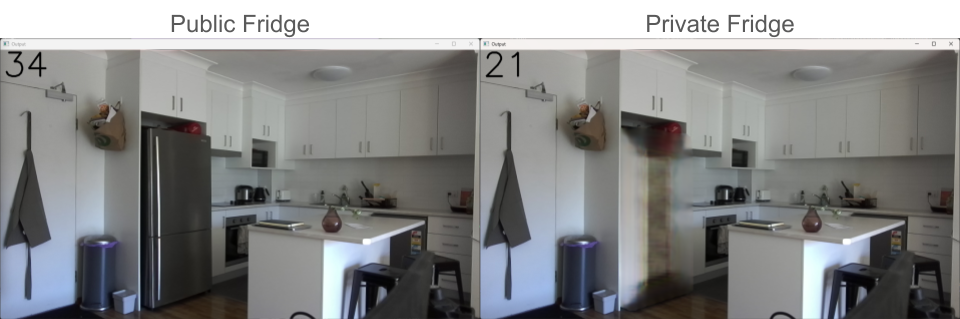}
    \caption{Central Visual Artefacts in Large Inpainted Regions}
    \label{fig:fridge}
\end{figure}

The model handles smaller regions extremely well, however larger regions produce noise towards the centre of the inpainting result. This means that users will likely notice larger inpainted regions existing within the frame, however the objects will still remain hidden, and the result will still be relatively unobtrusive in terms of colour and texture similarities. An example of this is seen in \autoref{fig:fridge} which has a noisy central area.

\subsubsection*{3D Point Cloud Render}
The privatised point cloud for remote 3D viewing is of high quality, and the room can be accurately viewed as a 3D space with private objects redacted.

\begin{figure}[htbp]
    \centering
    \includegraphics[width=\linewidth]{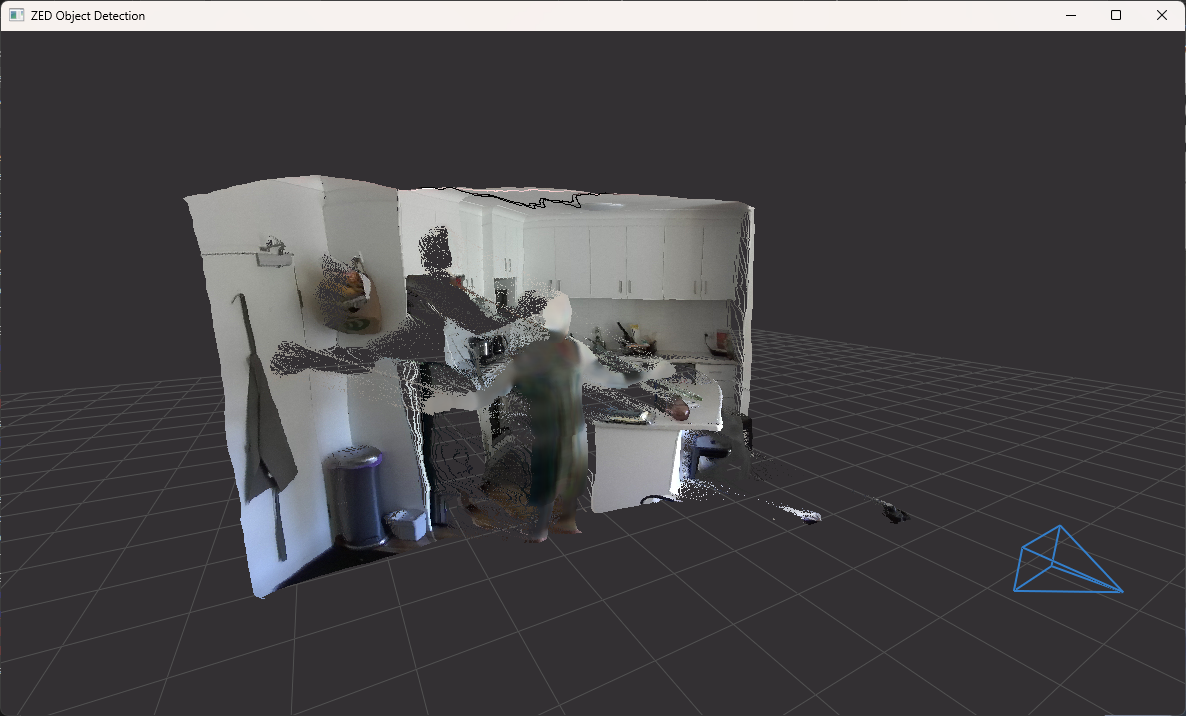}
    \caption{The Ghosting Effect}
    \label{fig:ghosting}
\end{figure}

Due to the inpainting being done on the 2D image alone, the depth map of the scene is not modified, and thus the inpainted pixels contain the same depth values. This results in an inpainted 'ghost' of the original object being produced, as seen in \autoref{fig:ghosting}. This does result in the contamination of the scene despite successful redaction. The 3D ghosting effect is less prominent for items which reside in non-central locations such as shelves or walls, seemingly fading into the surrounding material of similar depth. However, for items that protrude out into central space, the inpainted pixels are more noticeable and also become misaligned with the surrounding information when viewed at different angles. In these cases, the best results are found when viewing the scene from the same general angle as the ZED camera.

\section{Limitations}
The key limitations of the system are outlined as follows.
\subsection{Object Visualisation and Selection}
The current implementation only allows the removal of objects found by the YOLO segmentation model. Because the model is trained on the COCO dataset, only 80 objects can be recognised by it. This means that not all objects in a room can be detected, and therefore removed by the system.

Another downside of model-based object detection is that it is somewhat non-deterministic in nature. As a result, the outputs are somewhat inconsistent frame-to frame. This results in flickering as the model's certainty oscillates from high to low, meaning that objects can pop in and out of view. The object tracking provided by the ZED SDK is also not completely bullet proof. If a selected/hidden object is temporarily moved out of frame or another object occludes it briefly, tracking is often lost, and the object will return to a public state.

The positioning of bounding boxes is somewhat inaccurate. The ZED SDK ingests the 2D bounding boxes of the objects found by YOLO and calculates the depth by averaging the various depth readings from pixels in that region. In the event that the object is partially occluded by a closer object, the average depth values read will be offset by the other object within the 2D region. As a result, the bounding box may be positioned incorrectly. This artefact can be seen in \autoref{fig:flow-demo}, where the oven's 3D bounding box is positioned incorrectly due to the table occluding a large amount of area in its 2D bounding box.

\subsection{Depth Inpainting}
The model does not update the depth-map to reflect the redaction of objects. The modified pixels are transplanted onto the original depth map which means that the object's 'ghost' is still there, with hallucinated content placed on its surface. As a result of this, remote viewers will recognise that an object is indeed there, however the texture of the item will indeed be occluded. A more impressive system would also modify the perceived depth of the inpainted region to match surrounding data. This would produce a more visually consistent result.

\chapter{Discussion}

By drawing from the evaluation of the system's performance, we can outline any potential areas for improvement and directions of future work. By understanding the boundaries in which the system operates, we can recognise the implications of the project for both mixed reality privacy and real-time diminished reality applications.

\section{Future Work}

\section{User Interface}
The UI is relatively elementary and simplistic and does not offer a comprehensive set of interactions for the user. The system offers the essential tap-to-remove approach, however this could certainly be extended to include a variety of other privacy settings and interfaces. Small visual changes could also be made. For example, instead of bounding boxes being visualised, precise object outlines could be displayed, which would take up less physical space and provide a more obvious indication of which object is being selected.

\subsection{Further Performance Optimisations}
A system rebuild could be undertaken to keep as much of the data processing as possible on the GPU, rather than moving data from GPU to CPU incurring significant overhead. Various functions could likely be optimised to be more efficient, and certain operations could be converted to GPU kernels to be run on the GPU instead of the CPU. As recognised in the performance evaluation of the system, pre- and post-processing takes up a large portion of frame time. Any improvements to the data processing flow could greatly improve frame rates.

A smaller YOLO model could be used to reduce frame time spent detecting objects, as it currently is the most expensive component in the system.

Moving both the inpainting and segmentation models into the same process could produce a more efficient pipeline, however no solution was found after considerable research and testing. If a solution could be found, the pipeline could sustain higher frame rates.

With further performance optimisations, it may be possible to run the pipeline on the headset itself, allowing for a more portable privacy system without the need of edge computing.

\subsection{More Robust Object Tracking}
A more robust object tracking system could be developed to ensure that objects remain hidden once selected, and the fluctuating outputs from the object detection model can be tempered. Hybrid approaches combining anchor-based tracking with semantic segmentation such as the system designed by \cite{tabet_2024_utilityprivacy} would produce a far more reliable system. If this were coupled with a more comprehensive user interface system, any items or regions could be accurately and securely removed in a far more reliable manner.

\subsection{Depth Inpainting}
As MR becomes increasingly depth-aware, such models will likely be critical for delivering fully coherent diminished reality experiences. Various inpainting models have recently been developed to utilise depth readings for the inpainting process, also modifying the resulting depth mapping to match background geometry. This approach was not explored in this paper as it seems that the various models are not currently capable of real-time inpainting, however with various optimisations a more comprehensive depth-based system could be created. This would allow users to share their privatised spaces without the object 'ghosting' as explained earlier. Models which could be modified for real-time application could include DynaFill \cite{bei2020dynamic} or DeepDR \cite{gsaxner2023deepdrdeepstructureawarergbd}.

\section{Implications}
By evaluating the boundaries of the systems capabilities, we can recognise that this paper serves as a launchpad for potential future work, whilst still establishing a substantial baseline in the field of MR privacy solutions. It has been proven that diminished reality can be a powerful tool in the building of real-time privacy applications, and this work provides a functional foundation for its use in future MR privacy projects.
\chapter{Conclusion\label{cha:conclusion}}

With the growth of mixed-reality development and its integration into everyday life, headsets will be used in bedrooms, living rooms, and enterprise meeting rooms. Yet their success will ultimately depend on the trust of users that these devices will protect their privacy in such intimate spaces. The system presented in this thesis shows, for the first time, that fine-grained, object-level privacy control can be delivered in real time with only a single mobile depth camera and readily available GPU hardware.

The practical implication is straightforward but profound: MR developers no longer need multi-camera rigs or pre-scanned environment meshes to give users agency over what is visible to collaborators. Instead, they can drop-in a lightweight Diminished Reality (DR) service that keeps pace with real-time (\(\geq\)20 fps) expectations for head-worn displays, making fine-grained privacy control a default capability rather than an add-on.

Beyond immediate engineering convenience, the work contributes three strategic benefits:
\begin{itemize}
    \item Trust-by-design blueprint. By providing clear, colour-coded feedback in the headset UI, the system leverages long-standing XR privacy design principles to create an intuitive implementation that enables simple tap-to-remove controls. Placing the privacy tools in the users' hands builds trust, allowing for the technology to be adopted without friction.
    
    \item Deployment ready. Fine-grained redaction aligns with user expectations and could help developers incorporate privacy-preserving systems into varied applications without sacrificing immersion. The pipeline could also be applied to various system configurations, including object removal for mobile perspectives within a 3D rendered setting.

    \item Research springboard. The pipeline establishes a performance baseline against which future work on depth-aware inpainting, object tracking, or performance acceleration can be measured. This turns privacy tooling into an active area of development rather than leaving it as a hypothesised possibility.
\end{itemize}

Taken together, these contributions move privacy control from the hypothetical to the centre of MR experience design. They suggest a near-term future where inviting someone into your virtual workspace is as safe and socially frictionless as sharing a screen on today’s video-calling platforms.


\bibliographystyle{style/mybibstyle}
\bibliography{thesis}

\begin{thebibliography}{}

\bibitem[\protect\citename{B \bgroup \etal\egroup }2024]{tensor_gopik_2024}
Gopiktrishna~P B, Rubell Marion Lincy~G, Abhishek Rishekeeshan, and Deekshitha.
\newblock 2024.
\newblock Accelerating native inference model performance in edge devices using tensorrt.
\newblock In {\em 2024 IEEE Recent Advances in Intelligent Computational Systems (RAICS)}, pages 1--7.

\bibitem[\protect\citename{Bešić and Valada}2020]{bei2020dynamic}
Borna Bešić and Abhinav Valada.
\newblock 2020.
\newblock Dynamic object removal and spatio-temporal rgb-d inpainting via geometry-aware adversarial learning.
\newblock {\em arXiv preprint arXiv:2008.05058}.

\bibitem[\protect\citename{Butz \bgroup \etal\egroup }1998]{butz1998vampire}
Andreas Butz, Clifford Beshers, and Steven Feiner.
\newblock 1998.
\newblock Of vampire mirrors and privacy lamps: Privacy management in multi-user augmented environments.
\newblock In {\em Proceedings of the 11th Annual ACM Symposium on User Interface Software and Technology (UIST '98)}, pages 171--172, San Francisco, CA, USA, November. ACM.
\newblock Dept. of Computer Science, Columbia University, \texttt{\{butz,beshers,feiner\}@cs.columbia.edu}.

\bibitem[\protect\citename{Cheng \bgroup \etal\egroup }2022]{cheng_2022_towards}
Yi~Fei Cheng, Hang Yin, Yukang Yan, Jan Gugenheimer, and David Lindlbauer.
\newblock 2022.
\newblock Towards understanding diminished reality.
\newblock {\em CHI Conference on Human Factors in Computing Systems}, 04.

\bibitem[\protect\citename{De~Guzman \bgroup \etal\egroup }2019]{deguzman_2019_security}
Jaybie~A. De~Guzman, Kanchana Thilakarathna, and Aruna Seneviratne.
\newblock 2019.
\newblock Security and privacy approaches in mixed reality.
\newblock {\em ACM Computing Surveys}, 52:1--37, 10.

\bibitem[\protect\citename{Girshick \bgroup \etal\egroup }2016]{girshick_2016_rcnn}
Ross Girshick, Jeff Donahue, Trevor Darrell, and Jitendra Malik.
\newblock 2016.
\newblock Region-based convolutional networks for accurate object detection and segmentation.
\newblock {\em IEEE Transactions on Pattern Analysis and Machine Intelligence}, 38(1):142--158.

\bibitem[\protect\citename{Gsaxner \bgroup \etal\egroup }2023]{gsaxner2023deepdrdeepstructureawarergbd}
Christina Gsaxner, Shohei Mori, Dieter Schmalstieg, Jan Egger, Gerhard Paar, Werner Bailer, and Denis Kalkofen.
\newblock 2023.
\newblock Deepdr: Deep structure-aware rgb-d inpainting for diminished reality.

\bibitem[\protect\citename{Hacohen and Weinshall}2019]{hacohen2019powercurriculumlearningtraining}
Guy Hacohen and Daphna Weinshall.
\newblock 2019.
\newblock On the power of curriculum learning in training deep networks.

\bibitem[\protect\citename{Herling and Broll}2010]{herling2010advanced}
Jan Herling and Wolfgang Broll.
\newblock 2010.
\newblock Advanced self-contained object removal for realizing real-time diminished reality in unconstrained environments.
\newblock In {\em Proceedings of the 9th IEEE International Symposium on Mixed and Augmented Reality (ISMAR 2010)}. IEEE.

\bibitem[\protect\citename{Hill \bgroup \etal\egroup }2016]{hill_2016_on}
Steven Hill, Zhimin Zhou, Lawrence Saul, and Hovav Shacham.
\newblock 2016.
\newblock On the (in)effectiveness of mosaicing and blurring as tools for document redaction.
\newblock {\em Proceedings on Privacy Enhancing Technologies}, 2016:403--417, 07.

\bibitem[\protect\citename{Hornsey and Hibbard}2024]{Hornsey2024May}
Rebecca~L. Hornsey and Paul~B. Hibbard.
\newblock 2024.
\newblock {Current Perceptions of Virtual Reality Technology}.
\newblock {\em Appl. Sci.}, 14(10):4222, May.

\bibitem[\protect\citename{Jana \bgroup \etal\egroup }2013]{jana2013finegrained}
Suman Jana, David Molnar, Alexander Moshchuk, Alan Dunn, Benjamin Livshits, Helen~J. Wang, and Eyal Ofek.
\newblock 2013.
\newblock Enabling fine-grained permissions for augmented reality applications with recognizers.
\newblock In {\em Proceedings of the 22nd USENIX Security Symposium}, pages 415--430, Washington, D.C., USA, August. USENIX Association.

\bibitem[\protect\citename{Kato \bgroup \etal\egroup }2022]{kato2022online}
Taiki Kato, Naoya Isoyama, Norihiko Kawai, Hideaki Uchiyama, Nobuchika Sakata, and Kiyoshi Kiyokawa.
\newblock 2022.
\newblock Online adaptive integration of observation and inpainting for diminished reality with online surface reconstruction.
\newblock In {\em Proceedings of the IEEE International Symposium on Mixed and Augmented Reality Adjunct (ISMAR-Adjunct)}, pages 308--314. IEEE.

\bibitem[\protect\citename{Kobayashi and Takahashi}2024]{Kobayashi2024Mar}
Kaito Kobayashi and Masanobu Takahashi.
\newblock 2024.
\newblock {Real-Time Diminished Reality Application Specifying Target Based on 3D Region}.
\newblock {\em Virtual Worlds}, 3(1):115--134, March.

\bibitem[\protect\citename{Lebeck \bgroup \etal\egroup }2018]{lebeck2018security}
Kiron Lebeck, Kimberly Ruth, Tadayoshi Kohno, and Franziska Roesner.
\newblock 2018.
\newblock Towards security and privacy for multi-user augmented reality: Foundations with end users.
\newblock In {\em 2018 IEEE Symposium on Security and Privacy (SP)}, San Francisco, CA, USA. IEEE.
\newblock Paul G. Allen School of Computer Science \& Engineering, University of Washington.

\bibitem[\protect\citename{Li \bgroup \etal\egroup }2022]{li2022e2fgvi}
Zhen Li, Cheng-Ze Lu, Jianhua Qin, Chun-Le Guo, and Ming-Ming Cheng.
\newblock 2022.
\newblock Towards an end-to-end framework for flow-guided video inpainting.
\newblock {\em arXiv preprint arXiv:2204.02663}.
\newblock Accepted to CVPR 2022.

\bibitem[\protect\citename{Liu \bgroup \etal\egroup }2021]{Liu_2021_DSTT}
Rui Liu, Hanming Deng, Yangyi Huang, Xiaoyu Shi, Lewei Lu, Wenxiu Sun, Xiaogang Wang, and Li~Hongsheng.
\newblock 2021.
\newblock Decoupled spatial-temporal transformer for video inpainting.
\newblock {\em arXiv preprint arXiv:2104.06637}.

\bibitem[\protect\citename{Pathak \bgroup \etal\egroup }2016]{pathak_2016_context}
Deepak Pathak, Philipp Krahenbuhl, Jeff Donahue, Trevor Darrell, and Alexei~A Efros.
\newblock 2016.
\newblock Context encoders: Feature learning by inpainting.

\bibitem[\protect\citename{Perazzi \bgroup \etal\egroup }2016]{davis2016}
F.~Perazzi, J.~Pont-Tuset, B.~McWilliams, L.~Van~Gool, M.~Gross, and A.~Sorkine-Hornung.
\newblock 2016.
\newblock A benchmark dataset and evaluation methodology for video object segmentation.
\newblock In {\em 2016 IEEE Conference on Computer Vision and Pattern Recognition (CVPR)}, pages 724--732.

\bibitem[\protect\citename{Qian and Li}2022]{Qian2022Jun}
Feng Qian and Bin Li.
\newblock 2022.
\newblock {Boosting remote multi-user AR privacy through a magic rope}.
\newblock In {\em {ACM Conferences}}, pages 583--584. Association for Computing Machinery, New York, NY, USA, June.

\bibitem[\protect\citename{Quattoni and Torralba}2009]{quattoni2009indoor}
Ariadna Quattoni and Antonio Torralba.
\newblock 2009.
\newblock Recognizing indoor scenes.
\newblock In {\em Proceedings of the IEEE Conference on Computer Vision and Pattern Recognition (CVPR)}, pages 413--420, Miami, FL, USA. IEEE.

\bibitem[\protect\citename{Redmon \bgroup \etal\egroup }2016]{redmon2016yolo}
Joseph Redmon, Santosh Divvala, Ross Girshick, and Ali Farhadi.
\newblock 2016.
\newblock You only look once: Unified, real-time object detection.
\newblock {\em arXiv preprint arXiv:1506.02640}.
\newblock arXiv:1506.02640v5 [cs.CV], last revised 9 May 2016.

\bibitem[\protect\citename{Reilly \bgroup \etal\egroup }2014]{reilly2014secspace}
Derek Reilly, Mohamad Salimian, Bonnie MacKay, Niels Mathiasen, W.~Keith Edwards, and Juliano Franz.
\newblock 2014.
\newblock {SecSpace: Prototyping Usable Privacy and Security for Mixed Reality Collaborative Environments}.
\newblock In {\em Proceedings of the 2014 ACM SIGCHI Symposium on Engineering Interactive Computing Systems (EICS '14)}, pages 273--282. Association for Computing Machinery.

\bibitem[\protect\citename{Richter \bgroup \etal\egroup }2016]{richter2016playingdatagroundtruth}
Stephan~R. Richter, Vibhav Vineet, Stefan Roth, and Vladlen Koltun.
\newblock 2016.
\newblock Playing for data: Ground truth from computer games.

\bibitem[\protect\citename{Ruth \bgroup \etal\egroup }2019]{Ruth2019}
Kimberly Ruth, Tadayoshi Kohno, and Franziska Roesner.
\newblock 2019.
\newblock {Secure {$\lbrace$}Multi-User{$\rbrace$} Content Sharing for Augmented Reality Applications}.
\newblock [Online; accessed 21. May 2025].

\bibitem[\protect\citename{Tabet \bgroup \etal\egroup }2024a]{tabet_2024_blur}
Salam Tabet, Ayman Kayssi, and Imad~H. Elhajj.
\newblock 2024a.
\newblock Adaptive mobile diminished reality framework for 3d visual privacy.
\newblock In {\em 2024 International Conference on Intelligent Systems and Computer Vision (ISCV)}, pages 1--8.

\bibitem[\protect\citename{Tabet \bgroup \etal\egroup }2024b]{tabet_2024_utilityprivacy}
Salam Tabet, Ayman Kayssi, and Imad~H. Elhajj.
\newblock 2024b.
\newblock Utility-privacy aware mobile diminished reality framework for 3d visual privacy.
\newblock {\em 2024 IEEE 6th International Conference on Trust, Privacy and Security in Intelligent Systems, and Applications (TPS-ISA)}, pages 41--48, 10.

\bibitem[\protect\citename{Thiry \bgroup \etal\egroup }2024]{thiry2024memory}
Guillaume Thiry, Hao Tang, Radu Timofte, and Luc~Van Gool.
\newblock 2024.
\newblock Towards online real-time memory-based video inpainting transformers.
\newblock {\em arXiv preprint arXiv:2403.16161}.
\newblock arXiv:2403.16161v1 [cs.CV], 24 Mar 2024.

\bibitem[\protect\citename{Tian \bgroup \etal\egroup }2024]{tian2024yolov12}
Yunjie Tian, Qixiang Ye, and David Doermann.
\newblock 2024.
\newblock Yolov12: Attention-centric real-time object detectors.
\newblock {\em arXiv preprint arXiv:2502.12524}.
\newblock arXiv:2502.12524v1 [cs.CV], submitted on 22 Feb 2024.

\bibitem[\protect\citename{Xiao \bgroup \etal\egroup }2013]{xiao_sun3d_2013}
Jianxiong Xiao, Andrew Owens, and Antonio Torralba.
\newblock 2013.
\newblock Sun3d: A database of big spaces reconstructed using sfm and object labels.
\newblock In {\em Proceedings of 14th IEEE International Conference on Computer Vision (ICCV2013)}, pages 1625--1632, 12.

\bibitem[\protect\citename{Zhang}2020]{zhangsickness2020}
Chen Zhang.
\newblock 2020.
\newblock Investigation on motion sickness in virtual reality environment from the perspective of user experience.
\newblock In {\em 2020 IEEE 3rd International Conference on Information Systems and Computer Aided Education (ICISCAE)}, pages 393--396.

\bibitem[\protect\citename{Zhang}2024]{Zhang2024Dec}
Shengkun Zhang.
\newblock 2024.
\newblock {Research Advanced in Image Inpainting based on Deep Learning}.
\newblock {\em Highlights in Science Engineering and Technology}, 119:500--508, December.

\end{thebibliography}

\appendix
\chapter{Multi-Model GPU Runtime Conflicts} \label{appendix:multi-model}
Difficulties were faced when attempting to run both the YOLO segmentation engine and the inpainting engine within the same process. When both models were loaded within the same process, only the first model would produce valid outputs, while the second consistently produced incorrect results, often appearing as uninitialized or corrupted data. Several hypotheses were tested to isolate the root cause:
\begin{itemize}
    \item Ensured complete separation of memory buffers to prevent race conditions.
    \item Evaluated resource exhaustion, but ruled this out as both models ran simultaneously as standalone applications without exceeding VRAM limits.
    \item Unified both models to use FP32 precision to test for precision mismatch conflicts.
    \item Isolated CUDA contexts and streams, which was suspected to be the primary cause but ultimately did not resolve the issue.
\end{itemize}
Despite these efforts, concurrent execution remained unreliable.

As a result, the inpainting server was transferred into its own application and process, and IPC was utilised to efficiently pass content between the two.

\chapter{Inpainting Model Selection} \label{appendix:inpaint-choices}

One of the most critical steps in achieving real-time inpainting performance was converting the selected inpainting model into a TensorRT-compatible format. TensorRT, NVIDIA’s high-performance deep learning inference optimizer, requires models to be exported to the ONNX format with specific constraints; namely static input shapes, well-defined control flow, and limited dynamic indexing or slicing operations. This requirement posed a significant challenge due to the inherently dynamic structure of modern video inpainting models.

\section*{Initial Attempts with E2FGVI}
The first model evaluated was E$^{2}$FGVI, which is designed for efficient frame-guided video inpainting. Despite its impressive reconstruction quality, early attempts to export this model to ONNX quickly revealed major incompatibilities. The core issue stemmed from its dynamic control flow and use of unsupported operations. For example, the model relies on \texttt{aten::col2im}, which is unsupported in ONNX opset 16; the highest opset commonly supported by TensorRT tooling. While opset 18 added support for this operation, attempting to use \texttt{torch.onnx.dynamo\_export()} to leverage this newer opset resulted in further problems such as deep recursion issues within NumPy and unsupported dynamic slicing patterns.

Ultimately, the complexity and flexibility of E$^{2}$FGVI's structure made it infeasible to statically export without rewriting significant portions of the model. This led to the decision to pivot to a more static and reliable architecture: DSTT.

\section*{DSTT Conversion and Modification}
DSTT (Decoupled Spatial-Temporal Transformer) was chosen due to its lighter footprint, real-time inference aspirations, and hybrid CNN-transformer design. Even DSTT, which is inherently simpler than E$^{2}$FGVI, still required significant structural modifications to accommodate ONNX's and TensorRT’s strict requirements.

Many operations in the original PyTorch implementation were highly dynamic and Pythonic, which violated ONNX’s need for a statically-defined computation graph. Tensor shape calculations needed to be fixed, and thus various dynamically sized tensors caused issues as they changed throughout the inpainting process. Tracer warnings were also raised when Python floats or control flow were mixed with tensor computations making the traced model brittle and prone to incorrect generalization. 

PyTorch's \texttt{dynamo\_export()} improves support for Python control flow, but only when conditions are tensor-driven. Conditions based on Python variables were frozen as constants, harming model functionality. In order to produce a consistent model, the structural flow had to be solidified with consistent sizing and tensor-driven operations such that exportation could be completed.

Restructuring changes mentioned in \autoref{DSTT-mods} ultimately enabled the successful export of the model to ONNX (opset 17) and subsequent conversion to a TensorRT engine with FP16 precision. Despite the effort required to strip away some of the model’s flexibility and dynamic capacity, these trade-offs were essential in achieving the performance necessary for real-time mixed reality inpainting.

\chapter{Source Code} \label{appendix:sourcecode}

The inpainting management codebase can be found here:

\url{https://github.com/c1h1r1i1s1/inpaint_manager}

The main processing server can be found here:

\url{https://github.com/c1h1r1i1s1/AIO_server}

The modified inpainting model and training scripts can be found here:

\url{https://github.com/c1h1r1i1s1/Real-Time-Inpainting}

And finally, the Unity codebase for the headset application can be found here:

\url{https://github.com/c1h1r1i1s1/thesis-headset}
\end{document}